\documentclass[11pt,a4paper]{article}
\usepackage[hyperref]{acl2021}
\usepackage{times}
\usepackage{latexsym}

\usepackage{times}
\usepackage{latexsym}

\usepackage{xcolor,colortbl}
\usepackage{amsmath}
\usepackage{tabularx}
\usepackage{graphicx}
\usepackage{booktabs}
\usepackage{caption}
\usepackage{subcaption}
\usepackage{amssymb}
\usepackage{pdfrender}
\usepackage{amsmath}
\usepackage{pifont}

\usepackage{microtype}

\newcommand{\LMI}{$LM_\mathcal{I}$\ } 
 
\newcommand{\anli}{$\alpha$NLI\ } 
\newcommand{\MTL}{$\mathcal{MTL}$\ }

\title{Generating Hypothetical Events for Abductive Inference}

\author{Debjit Paul \\
  Research Training Group AIPHES \\
  Institute for Computational Linguistics\\
  Heidelberg University \\
  {\tt paul@cl.uni-heidelberg.de} \\\And
  Anette Frank \\
  Research Training Group AIPHES \\
  Institute for Computational Linguistics \\
  Heidelberg University\\
  {\tt frank@cl.uni-heidelberg.de} \\}
\aclfinalcopy
\date{}

\begin{document}
\maketitle
\begin{abstract}

Abductive reasoning starts from some observations and aims at finding the most plausible explanation for these observations. To perform abduction, humans often make use of temporal and causal inferences, and knowledge about how some hypothetical situation can result in different outcomes. This work offers the first study of how such knowledge impacts the \textit{Abductive $\alpha$\textbf{NLI}} task -- which consists in choosing the more likely explanation for given observations. We train a specialized language model \LMI that is tasked to generate \textit{what could happen next} from a hypothetical scenario that evolves from a given event. We then propose a multi-task model \MTL to solve the \textbf{\anli task}, which  predicts a plausible explanation by a) considering different \textit{possible events} emerging from candidate hypotheses -- events generated by \LMI -- and b) selecting the one that is most  \textit{similar} to the observed outcome. We show that our \MTL model improves over prior vanilla pre-trained LMs fine-tuned on $\alpha$NLI. Our manual evaluation and analysis suggest that learning about possible next events from different hypothetical scenarios supports abductive inference. 


\end{abstract}
\section{Introduction}
\begin{figure}
    \centering
    \includegraphics[scale=1.0,height=2.0cm, width=0.45\textwidth]{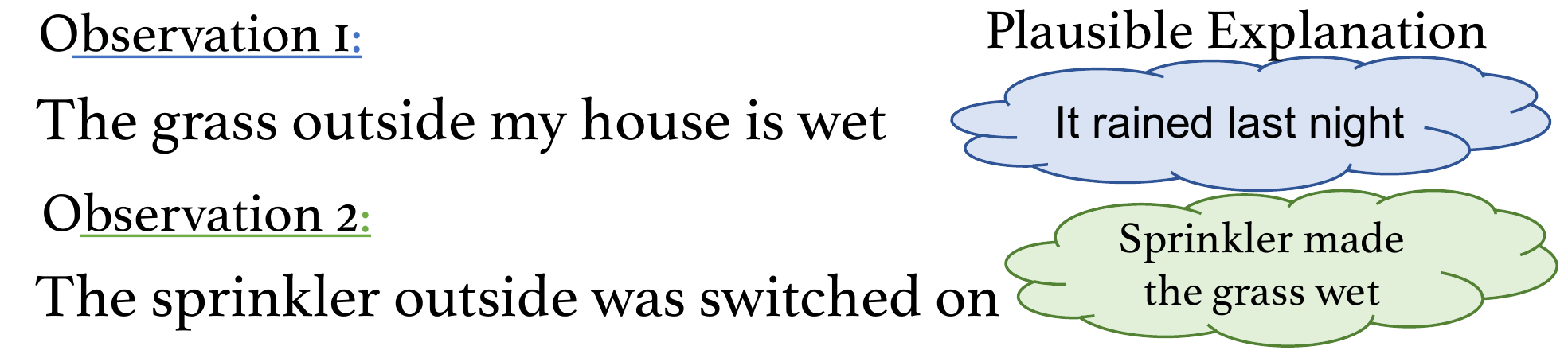}
    \caption{
    Motivational example 
    illustrating Abductive Reasoning and its non-monotonic character.} 
    \label{fig:example1}
\end{figure}

Abductive reasoning (AR) is inference to the best explanation. It typically 
starts from an incomplete set of observations about everyday situations and 
comes up with what can be considered
the most likely 
possible 
explanation given these observations \citep{pople1973mechanization, sep-abduction}. 
One of the key characteristics that make abductive reasoning more challenging and distinct from other types of reasoning is its non-monotonic character \cite{sep-logic-nonmonotonic} i.e., even the most likely explanations are not necessarily correct. For example, in Figure \ref{fig:example1}, the most likely explanation for \textit{Observation 1: ``wet grass outside my house''} is that \textit{``it has been raining''}. However, when a new piece of information (observation or evidence) becomes available, the explanation must possibly be retracted, \textit{showing the defeasible character of abduction}. With the new observation (\textit{``the sprinkler was switched on}'') the most plausible explanation changes to \textit{``Sprinkler caused the grass to be wet''}. Humans, in such situations, could induce or validate such abductive inferences by performing hypothetical reasoning (such as \textit{``What would happen if the sprinkler was switched on?''}) to arrive at a plausible explanation for \textit{``wet grass outside my house''}. 

\begin{figure}[t]
  \centering
    \includegraphics[scale=1.0,height=3.5cm, width=0.48\textwidth]{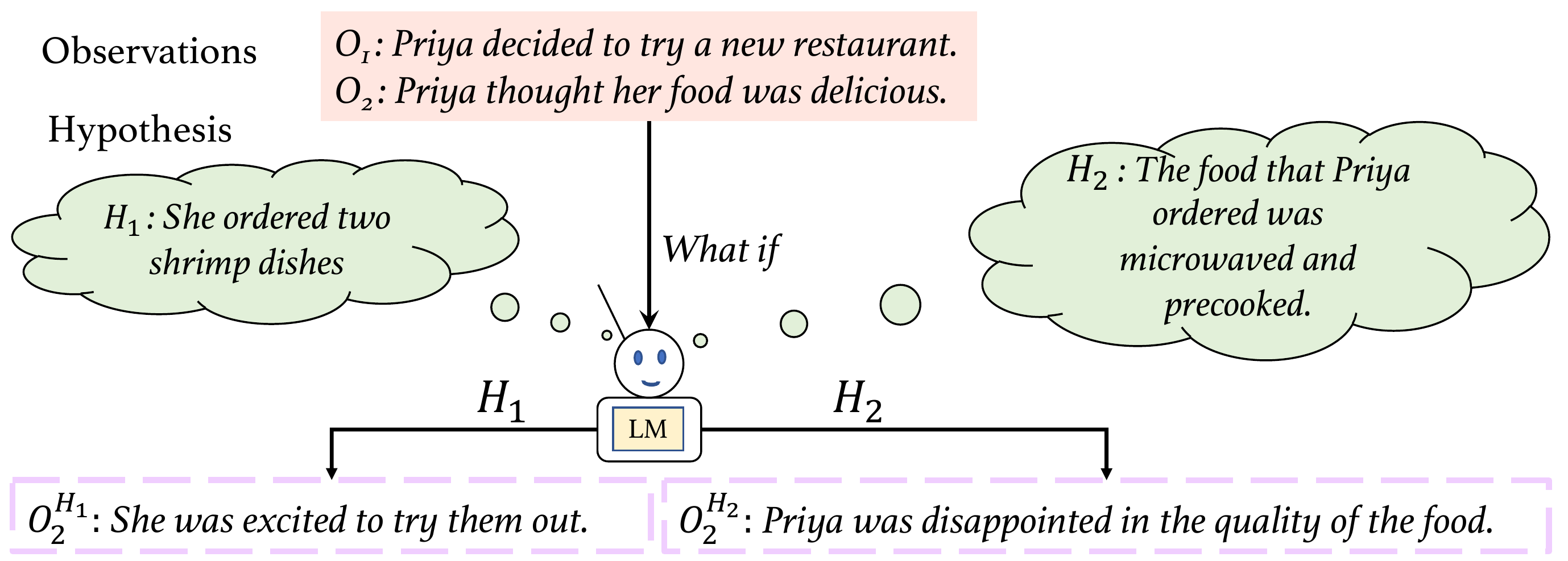}
    \caption{Motivational example for \anli: The top box (red) shows the 
    observations and two callout clouds (green) contain the hypotheses.  The 
    implications ($O_i^{H_i}$)
    -- generated by the LM conditioned on each hypothesis and 
    the observations 
    -- are given in pink colored boxes.
    }
    \label{fig:example2}
\end{figure}

In this work, we focus on the \textbf{\anli task} \cite{bhagavatula2019abductive}, where given two observations (\textit{$O_1$} at time $t_1$, \textit{$O_2$} at time $t_2$, with $t_1 < t_2$) as an incomplete context, the task is to predict which of two given hypothesized events ($H_1$ or $H_2$) is more plausible to have happened between \textit{$O_1$} and \textit{$O_2$}.
Figure \ref{fig:example2} illustrates this with an example: given 
observations \textit{$O_1$:``Priya decided to try a new restaurant.''} and \textit{$O_2$:  ``Priya thought her food was delicious.''}, the task is to predict whether \textit{$H_1$} or \textit{$H_2$} is the more plausible explanation given observations $O_1$ and $O_2$. 
Both \textit{$H_1$} and \textit{$H_2$} are different plausible hypothetical situations  
that can evolve from the same observation (premise) $O_1$.

In this paper, we hypothesize that learning how different hypothetical scenarios (\textit{$H_1$} and \textit{$H_2$}) can result in different outcomes (e.g., $O_2^{H_j}$, Fig.\ \ref{fig:example2}) can help in performing abductive inference. In order to decide which \textit{$H_i$}, is \textit{more plausible} given observations, we assume each $H_i$ to be \textit{true} and generate a \textit{possible next event $O_2^{H_i}$} for each of them independently (e.g.: \textit{What will happen if Priya's ordered food was microwaved and precooked?}). We then compare the generated sentences ($O_2^{H_1}$, $O_2^{H_2}$ in Fig.\ 2) to what has been observed ($O_2$) and choose as most plausible hypothesis the one whose implication is closest to observation $O_2$. 

We design a language model (\LMI) which, given  observations and a hypothesis, generates a possible event that could happen next, given one hypothesis. In order to train this language model,
we use the TIMETRAVEL (TT) corpus \cite{qin-counterfactual} (a subpart of the \textit{ROCStories} corpus\footnote{We ensure that $\alpha$NLI testing instances are held out.}).
We utilize the \LMI model to generate a possible next event for each hypothesis, given the observations. We then propose a multi-task learning model \MTL that jointly chooses from the generated possible next events ($O^{H_1}_{2}$ or $O^{H_2}_{2}$) the one most similar to the observation $O_2$ and predicts the most plausible hypothesis ($H_1$ or $H_2$). 


Our contributions are: i) To our best knowledge, we are the first to demonstrate that a model that learns to perform hypothetical reasoning can support and improve abductive tasks 
such as $\alpha$NLI. We show that ii) for $\alpha$NLI our multi-task model outperforms a strong BERT baseline \cite{bhagavatula2019abductive}. 

Our code is made publicly available.\footnote{\url{https://github.com/Heidelberg-NLP/HYPEVENTS}}



\if false
\begin{table*}[t]
\centering
\small
\scalebox{1.0}{
\begin{tabular}{@{}p{4cm}p{4cm}p{3cm}p{3cm}}
\toprule
\textbf{Observation} & \textbf{Hypothesis} &\textbf{FI} & \textbf{CFI}\\\hline
\textit{Priya decided to try a new restaurant.  Priya thought her food was delicious.
} & {\textbf{She ordered two shrimp dishes.}} & {priya ate her shrimp dish.} & {she was excited to try them out.} \\
{} & {The food that Priya ordered was microwaved and precooked.} & {priya was very happy with her new meal.} & {priya was disappointed in the quality of the food.} \\\hline
\textit{Jim got ready for his first date. Since then, she has ignored all of Jim's text messages.} & { \textbf{Jim's date wasn't attracted to him.}} & {Jim was disappointed.} & {He tried to get her number. She never responded to him.} \\
{} & {Jim went on the date and said he didn't like the girl.} & {The girl said she liked Jim} & {He told her that he didn't want to date her.} \\
\bottomrule
\end{tabular}
}
\caption{Example of Forward Inference.
 }
\label{tbl:examples}
\end{table*}
\fi 
\section{Learning about Counterfactual Scenarios}

\begin{figure}[t]
  \centering
    \includegraphics[scale=1.0,height=6cm, width=0.48\textwidth]{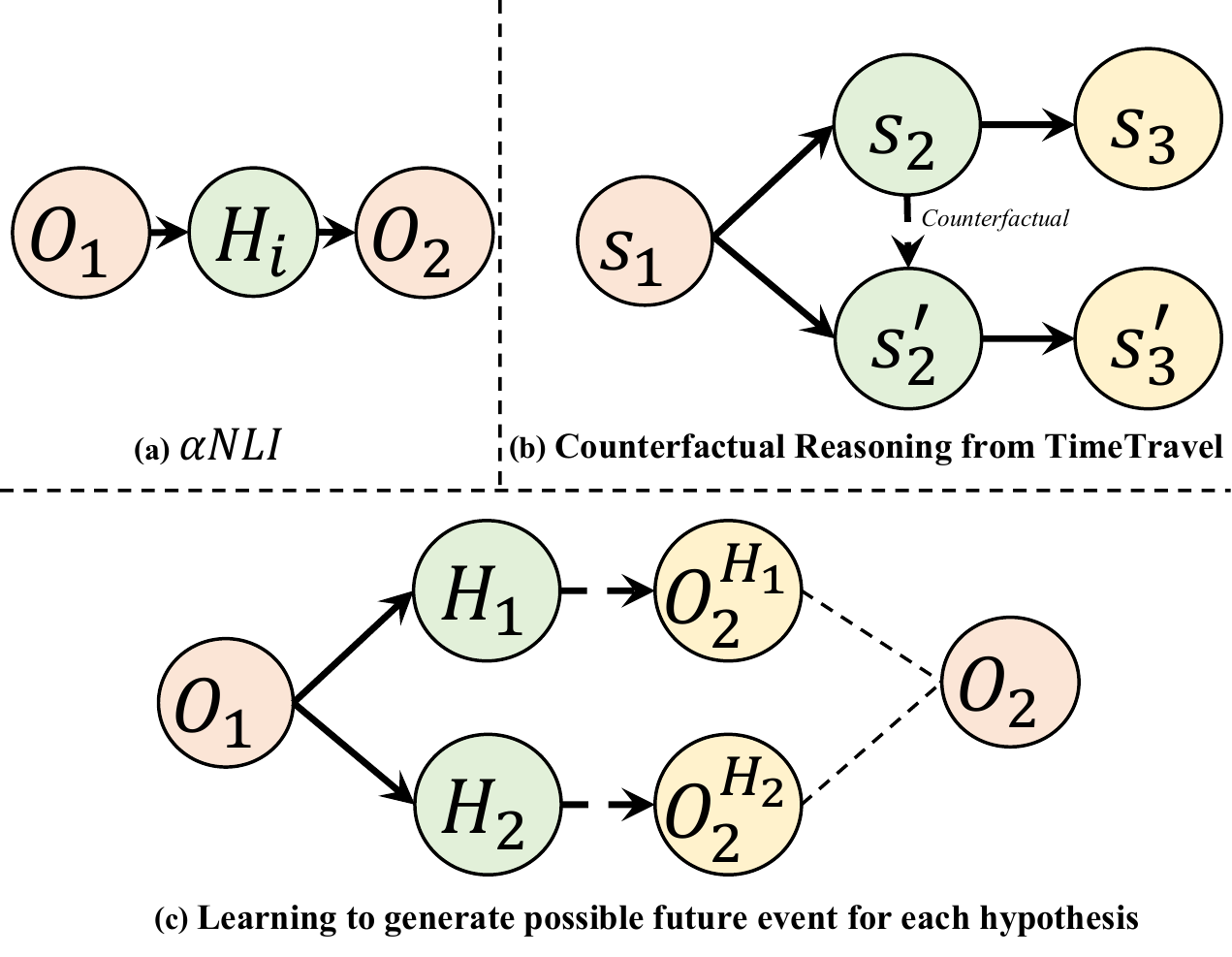}
    \caption{Different reasoning schemes and settings for our task and approach. The arrows denote the direction (temporal flow) of the reasoning chain. The dotted arrow in (b) denotes the derivation of a counterfactual situation $s'_2$ from a factual $s_2$. 
    In (c), the dotted arrows 
    denote 
    the learned inference; the dotted lines 
    indicate the similarity between $O_2$ and $O_{2}^{H_i}$.}
    \label{fig:causual_graph}
\end{figure}

The main idea is to learn to generate assumptions, in a given situation, about \textit{``What could have happened (next) if we had done X?'' or ``What could happen (next) if we do X?''} \cite{Bhatt2010SpatioTemporalAF}. 
Figure \ref{fig:causual_graph}(a) depicts the \anli task framework. We hypothesize that getting to know \textit{what will happen (next) if any of two hypotheses occurs}, will help us verifying which 
of 
them is more plausible (see Fig. \ref{fig:causual_graph}(c)). Therefore, we encourage the model to learn how different hypothetical events (including counterfactual events) evolving from the same premise ($s_1$) can lead to different or similar outcomes (see Fig. \ref{fig:causual_graph}(b)).

Accordingly, we teach a pre-trained GPT-2 \cite{radford2019language} language model how to generate \textit{a sequence of} possible subsequent events given different hypothetical situations in a narrative setting. Training such a model on narrative texts encourages it to learn causal and temporal relations between events. We train a conditional language model, \LMI, which generates a possible event that could happen next, given some
counterfactual scenarios for a given story.

We train this model on the TIMETRAVEL (TT) dataset \cite{qin-counterfactual},
by fine-tuning GPT-2 to learn about possible next events emerging from a situation in a story, given some alternative,
counterfactual 
event.
The TT dataset consists of five-sentence instances $S$=($s_1$,$s_2$,..,$s_5$)\footnote{$s_1$ = \textit{premise}, $s_{2}$ = \textit{initial context}, $s_{3:5}$ = original ending} from the ROCStories corpus\footnotemark[1]
plus additional crowd-sourced sentences ${s^{'}_{2:5}}$, where ${s^{'}_{2}}$ is counterfactual\footnote{a counterfactual $s^{'}$ states something that is contrary to $s$} to $s_{2}$ from the original story\footnote{During our experiments we treat them as two separate instances: ${S_1}$=(${s_{1:5}}$) and ${S_2}$ = (${s_1}$,${s^{'}_{2:5}}$).}.
There are two reasons for using the TT dataset 
for our purposes:
a) the domains on which GPT-2 was pretrained are broad\footnote{GPT-2 was trained on the WebText Corpus.} and different from the domain of ROCStories,
b) the model can see how alternative situations can occur starting from the same premise $s_1$, resulting
in similar or different outcomes. Note that, although intermediate 
situations may be
counterfactual to each other, the future outcome can still be similar to the original ending due to \textit{causal invariance} \footnote{the future events that are invariant under the counterfactual conditions \cite{qin-counterfactual}}.

Concretely, the language model \LMI reads the premise ($s_1$) and the alternative event(s) ($s_2$ or $s^{'}_2$), the masked token (serving
as a placeholder for the missing possible next event(s) ($s_{3:i}$ or $s^{'}_{3:i}$), 
then the rest of the story ($s_{i+1:5}$ or $s^{'}_{i+1:5}$) and again the premise ($s_1$).  We train the model to maximize the log-likelihood of the missing ground-truth sentence(s) ($s_{3:i}$). 

\vspace*{-1.5mm}
\begin{equation}
\begin{aligned}
    \mathcal{L}^{LM_\mathcal{I}}(\beta)=\\ log_{p_{\beta}}(s_{3:i}|[S] s_{1}, [M], s_{i+1:5}, [E], [S], s_{1}, s_{2}) \\
\hspace*{-3mm}    + log_{p_{\beta}}(s^{'}_{3:i}|[S] s_{1}, [M], s^{'}_{i+1:5}, [E], [S], s_{1}, s^{'}_{2})\label{eq:anlitrain}
\end{aligned}
\end{equation}
where ${i \in [3,4]}$, $s_i$=\{${w^{s_i}_1,..,w^{s_i}_n}$\} a
sequence of tokens, $[S]$=start-of-sentence token, $[E]$=end-of-sentence token, $[M]$=mask token.

\begin{table}[t]
\centering

\begin{tabular}{l}
\textit{$O_1$: Dotty was being very grumpy.} \\
\textit{$O_2$: She felt much better afterwards.}\\
{\textit{$H_1$}: Dotty ate something bad.} \\
{\textit{$H_2$}: \textbf{Dotty call some close friends to chat.}}\\
\textit{$O^{H_1}_{2}$: She started to feel sick.}  \\
\textit{$O^{H_2}_{2}$: They all tried to make her happy.} \\
\end{tabular}

\caption{Example of generated possible next events $O^{H_j}_{2}$ using the \LMI model. \textbf{Bold} hypothesis ($H_2$) is more plausible.}
\label{tab:fiexamples}
\end{table}


\section{Generating Hypothetical Events to support the $\alpha$NLI task}
In this paper, we aim to investigate whether 
models perform better on the $\alpha$NLI task when explicitly learning about events that could follow other events in a hypothetical scenario. We do so by introducing two methods $LM_\mathcal{I}$ + \textit{BERTScore} and $LM_\mathcal{I}$ + $\mathcal{MTL}$ for unsupervised and supervised settings, respectively. 

We first apply the trained model \LMI on the $\alpha$NLI task, where the given observations $O_1$ and $O_2$, and alternative hypotheses $H_j$ are fed as shown in (2) below.\footnote{For definition of placeholders see (\ref{eq:anlitrain}).}
\begin{equation}
\begin{aligned}
\hspace*{-3mm}  O^{H_j}_{2} = {\beta}([S],O_{1},[M], O_{2},[E],[S],O_{1}, H_{j})
    \label{eq:generate_next}
\end{aligned}
\end{equation}

We ge\-ne\-rate a possible next event for each hypothetical event $H_j$, i.e., $O^{H_1}_{2}$ and $O^{H_2}_{2}$ (or:  
what will happen if some hypothesis $H_j$ occurs given the observations), where $j\in$ $[1,2]$. Table \ref{tab:fiexamples} illustrates an example where different $O^{H_j}_{2}$ are generated using $LM_\mathcal{I}$. One of the challenges when generating subsequent events given a hypothetical situation is that there can be infinite numbers of possible next events. Therefore, to constrain this range, we chose to give future events ($O_2$) as input, such that the model can generate subsequent events in a constrained context.


\subsection{Unsupervised Setting}

In this setting, we do not train any supervised model to explicitly predict which hypothesis is more plausible given the observations. Instead, we apply the fine-tuned \LMI model to the $\alpha$NLI data, generate possible next events $O_2^{H_j}$ given $O_1$ and $H_j$, as described above, and measure the similarity between such possible next events ($O_2^{H_j}$) 
and the observation ($O_2$) in an unsupervised way, using \textit{BERTScore} (BS) \cite{Zhang2020BERTScore} \footnote{BERTScore is an automatic evaluation metric for text generation that leverages the pre-trained contextual embeddings from BERT and matches words in candidate and reference sentences by cosine similarity.}.
We evaluate our hypothesis that the generated possible next event $O^{H_j}_{2}$ given the more
plau\-si\-ble hypothesis $H_j$ should be \textit{more similar} to observation $O_{2}$. Table \ref{tab:fiexamples} illustrates an example where {$H_2$} is the more plausible hypothesis.
We impose the constraint that for a correctly predicted instance BS(${O_2}^{H^{+}}$, $O_2$) $>$ BS(${O_2}^{H^{-}}$, $O_2$) should hold, where ${H^{+}}$, ${H^{-}}$ are the more plausible vs. implausible hypothesis, respectively.

\subsection{Supervised Setting} 

\begin{figure}[t]
  \centering
    \includegraphics[scale=1.0,height=6.0cm]{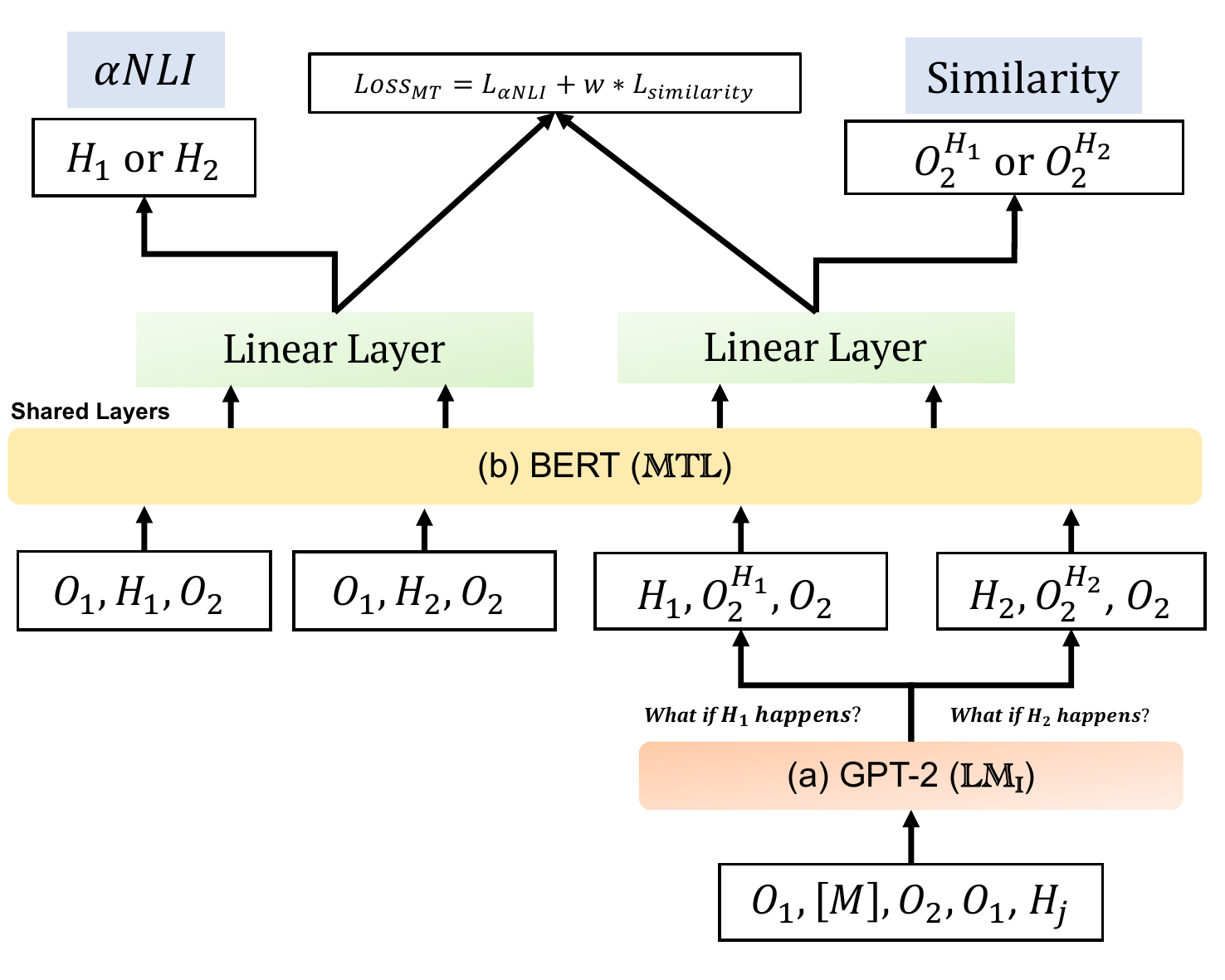}
    \caption{ 
    Overview of our \LMI+ $\mathcal{MTL}$ 
    model for $\alpha$NLI: (a) 
    language model \LMI takes the input in a particular format to generate different possible next events, (b) 
    the $\mathcal{MTL}$ model 
    learns to predict the best explanation ($H_j$) and possible next events ($O^{H_j}_{2}$) at the same time to perform the $\alpha$NLI task.}
    \label{fig:mtl}
\end{figure}

In this setting, displayed in Figure \ref{fig:mtl}, we explore the benefits of training a multi-task $\mathcal{MTL}$ model that predicts i) the most plausible hypothesis and ii) which possible next event ($O^{H_j}_{2}$) is more similar to the observation ($O_2$). Multi-task learning aims to improve the performance of a model for a task by utilizing the knowledge acquired by learning related tasks \cite{Ruder2017AnOO}. We \textit{hy\-po\-the\-si\-ze that} a) the  possible next event $O^{H_j}_{2}$ of the more plau\-si\-ble hypothesis $H_j$ should be most similar to  observation $O_{2}$, and that b) learning which possible next event is more similar supports the model in the $\alpha$NLI task (\textit{inductive transfer}).

The architecture of \LMI+ $\mathcal{MTL}$ model is shown in Figure \ref{fig:mtl}. The model marked (a) in Figure \ref{fig:mtl} depicts the \LMI model as described in \S 3. The outputs of the \LMI model, which we get from Eq. (\ref{eq:generate_next}) for both hypotheses are incorporated as an input to the $\mathcal{MTL}$ model.
Concretely, we feed the $\mathcal{MTL}$ classifier a sequence of tokens as stated in part (b) of Figure \ref{fig:mtl}, and aim to compute their contextualized representations using pre-trained BERT. The input format is described in Table \ref{tab:format}. Similar to \cite{devlin-etal-2019-bert}, two additional tokens are added [CLS] at the start of each sequence input and [SEP] at the end of each sentence. In the shared layers (see Fig \ref{fig:mtl}(b)), the model first transform the input sequence to a sequence of embedding vectors. Then it applies an attention mechanism that learns contextual relations between words (or sub-words) in the input sequence.

For each instance we get four [CLS] embeddings ($CLS_{H_{j}}, CLS_{O^{H_j}_{2}}$; $j \in [1,2]$) which are then passed through two linear layers, one for the $\alpha$NLI (main task) and another for predicting 
the similarity (auxiliary task) between $O^{H_j}_{2}$ and $O_2$. We compute the joint loss function $\mathcal{L}$ = $\mathcal{L}_{\alpha NLI}$ + $w*\mathcal{L}_{similarity}$; 
where $w$ is a trainable parameter, $\mathcal{L}_{\alpha NLI}$ and $\mathcal{L}_{similarity}$ are the loss function for the  $\alpha NLI$ task and auxiliary task, respectively.  


\section{Experimental Setup} 
\begin{table}[t!]
\centering{
\scalebox{0.9
}{
\begin{tabular}{@{}llll@{}}
\toprule
{\bf Task} & {\bf Train}& {\bf Dev} & {\bf Test}\\
\midrule
$\alpha$NLI & {169654}& {1532} & {3059} \\
TimeTravel (NLG) & {53806}& {2998} & {--} \\
\bottomrule
\end{tabular}}
\caption{Dataset Statistics:
nb.\ of instances}\label{tab:data_stat}}
\end{table}

\begin{table}[t!]
\scalebox{0.9}{
\centering
\begin{tabular}{ll}
\hline
{\bf Input Format} & {\bf Output} \\
\hline 
{[CLS] $O_1$ [SEP] $H_i$ [SEP] $O_2$ [SEP]} & {$H_1$ or $H_2$} \\
{[CLS] $H_i$ [SEP] $O^{H_i}_{2}$ [SEP] $O_2$[SEP]} & {$O^{H_1}_{2}$ or $O^{H_2}_{2}$}\\
\bottomrule
\end{tabular}}
\caption{Input and output format for the $\alpha$NLI task: [CLS] is a special token used for classification, [SEP] a delimiter.}\label{tab:format}
\end{table}

\textbf{Data.} We conduct 
experiments on the $\mathcal{ART}$ \cite{bhagavatula2019abductive} dataset. 
Data statistics are given in Table \ref{tab:data_stat}. For evaluation, we measure accuracy for $\alpha$NLI.

\paragraph{Hyperparameters.} 
To train the \textbf{\LMI} model we use learning rate of $5e-05$. We decay the learning rate linearly until the end of training; batch size: 12. In the supervised setting for the \textbf{$\alpha$NLI task}, we use the following set of hyperparameters for our $\mathcal{MTL}$ model with integrated \LMI model ($LM_\mathcal{I}+{\mathcal{MTL}}$): batch size: \{$8, 16$\}; epochs: \{$3, 5$\}; learning rate: \{$2e$-$5$, $5e$-$6$\}. For evaluation, we measure accuracy. We use Adam Optimizer, and dropout rate = $0.1$. We experimented on GPU size of $11$GB and $24$GB. Training is performed using cross-entropy loss. The loss function is $\mathcal{L}_{\alpha NLI}$ + $w*\mathcal{L}_{similarity}$, where $w$ is a trainable parameter. During our experiment we initialize $w=1$. The input format is depicted in Table \ref{tab:format}. We report performance by averaging results along with the variance obtained for 5 different seeds. 

\paragraph{Baselines.} We compare to the following
baseline models that we apply to the $\alpha$NLI task, training them
on the training portion of the $\mathcal{ART}$ dataset (cf.\ Table \ref{tab:data_stat}).


\begin{itemize}
    \item \textit{ESIM + ELMo} is based on the ESIM model previously used for NLI \cite{chen-etal-2017-enhanced}. We use (a) ELMo to encode the observations and hypothesis, followed by (b) an attention layer, (c) a local inference layer, and (d) another bi-directional LSTM inference composition layer, and (e) a pooling operation, 
    \item \textit{Infersent} \cite{conneau-etal-2017-supervised} uses sentence encoding based on a bi-directional LSTM architecture with max pooling.
    \item \textit{BERT} \cite{devlin-etal-2019-bert} is a LM trained with a masked-language modeling (MLM) and next sentence prediction objective. 
\end{itemize}

As baselines for using the $\mathcal{MTL}$ model, we replace \LMI  with alternative generative LMs:

\begin{itemize}
    \item \textit{GPT-2} + $\mathcal{MTL}$. In this setup, we directly use the pretrained GPT-2 model and task it to generate a next sentence conditioned on each hypothesis ($O^{H_i}_{2}$) without finetuning it on the TIMETRAVEL data. We then use the supervised $\mathcal{MTL}$ model to predict the most plausible hypothesis and which of the generated observations is more similar to $O_2$.

    \item \textit{COMET} + $\mathcal{MTL}$. In this setting, we make use of inferential \textit{if-then} knowledge from ATOMIC \cite{sapatomic} as background knowledge. Specifically, we use COMET to generate objects with \textbf{Effect}\footnote{as a result PersonX feels; as a result PersonX wants; PersonX then} relations for each hypothesis as a textual phrase.
\end{itemize}

\begin{table}[t!]
\centering{
\scalebox{0.89}{
\begin{tabular}{@{}l@{~}c@{~}c@{~}}
\toprule
{\bf Model} & {\bf Dev Acc.(\%)} & {\bf Test Acc.(\%)} \\\midrule
Majority (\textit{from dev set})$^\diamond$ & -- &{50.8}\\
\rowcolor{blue!15}
$LM_\mathcal{I}$ + BERTScore & 62.27& \textbf{60.08} \\\hline
Infersent $^\diamond$ &{50.9}& 50.8 \\
ESIM + ELMo $^\diamond$ &{58.2}& 58.8 \\
BERT$_{Large}$ $^\diamond$ &{69.1}& {68.9}$\pm{0.5}$ \\ 
GPT-2 $+ \mathcal{MTL}$ &{68.9}$\pm{0.3}$& 68.8$\pm{0.3}$ \\
COMET $+ \mathcal{MTL}$ &{69.4}$\pm{0.4}$& 69.1$\pm{0.5}$ \\
\rowcolor{blue!15}
$LM_\mathcal{I} + \mathcal{MTL}$ &{72.9}$\pm{0.5}$& \textbf{72.2}$\pm{0.6}$ \\
\hline 
\textit{Human Performance} &-& 91.4\\
\bottomrule
\end{tabular}}} \caption{Results on \textbf{\anli task}, ${\diamond}$ : as in \citet{bhagavatula2019abductive} {\footnotesize(no unpublished leaderboard results)}. For each row, the best results are in bold, and performance of our models are in blue.}\label{tab:abnli}
\end{table}

\begin{table*}[t!]
\centering
\scalebox{0.7}{
\begin{tabular}{@{}p{0.2cm}p{4cm}p{4.5cm}p{4.5cm}p{1.5cm}p{1.5cm}p{1cm}p{1.5cm}}
\toprule
{} & \textbf{Observation ($O$)} & \textbf{Given Hypotheses ($H_j$)} & \textbf{What \textit{if} $H_j$ happens?} \newline  \textbf{Ge\-ne\-rated possible next events $O_{2}^{H_j}$} & \textbf{MTL} & \textbf{Reasoning Type} & \textbf{BERT Score} & \textbf{Human Evaluation}\\\hline
{\textbf{a.}} &${O_1}$: Jim got ready for his first date. & { $H_1$: \textbf{ Jim's date wasn't attracted to him.}} & {$O_{1}^{H_2}$: He tried to get her number. She never responded to him.} & $H_1$,  $O_{2}^{H_1}$& \textit{Reaction} & \cellcolor{green!15}{\textbf{47.34}} &  Non-Contradiction \\
{}& ${O_2}$: Since then, she has ignored all of Jim's text messages. & {$H_2$: Jim went on the date and said he didn't like the girl.} & {$O_{2}^{H_2}$: He told her that he didn't want to date her.}&&&{44.19} & Contradiction\\\hline
{\textbf{b.}} & ${O_1}$: Nate had the summer off before college. & $H_1$: {\textbf{Nate spent the summer traveling and partying.}} & $O_{2}^{H_1}$: {Nate had a lot of fun.} &$H_1$,  $O_{2}^{H_1}$&\textit{Motivation}& \cellcolor{green!15}{\textbf{49.02}} & Non-Contradiction \\
{}&${O_2}$: Nate's last summer before college was a total blast! & $H_2$:{Nate decided to spend the entire summer working in the Mines.} & $O_{2}^{H_2}$: {Nate spent the summer working in the mines.} &&& {44.34} & Redundancy\\\hline
{\textbf{c.}} & ${O_1}$: Dotty was being very grumpy. & $H_1$: {Dotty ate something bad.} & $O_{2}^{H_1}$: {She started to feel sick.} &$H_2$,  $O_{2}^{H_2}$&\textit{Emotional Effect}& \cellcolor{red!15}\textbf{0.604} & Contradiction\\
{}&${O_2}$: She felt much better afterwards & $H_2$: {\textbf{Dotty call some close friends to chat.}} & $O_{2}^{H_2}$: {They all tried to make her happy.} &&& {0.405} & Non-Contradiction\\\hline
{\textbf{d.}} & ${O_1}$: Cay had a crush on a boy in her class.	 & $H_1$: \textbf{Cay sent a love note to the boy.} & $O_{2}^{H_1}$: {The boy responded and said he liked Cay.} &$H_1$, $O_{2}^{H_1}$&\textit{Emotional Effect}& \cellcolor{green!15}\textbf{0.509} & Non-Contradiction\\
{}&${O_2}$: He smiled at her after and said he liked her too! & $H_2$: {She told him she did not like him.} & $O_{2}^{H_2}$: {The boy was very sad about it.} &&& \cellcolor{red!15}{0.423} & Contradiction\\
\bottomrule
\end{tabular}}\caption{Examples of generated possible next events for solving $\alpha$NLI using our \LMI model. Column 3: Hypothesis and possible next events chosen by our $LM_\mathcal{I}+\mathcal{MTL}$ model; Column 4: Reasoning type between the hypothesis ${H_j}$ and $O_2$; Column 5: BERTScore between the ${O_2^{H_j}}$ and $O_2$; Column5: Human evaluation of the possible next events with respect the observation $O_2$.} 
\label{tab:anli_examples}
\end{table*}

\section{Results} 


In Table \ref{tab:abnli}, we compare our models $LM_\mathcal{I}$ + BERTScore and $LM_\mathcal{I}+ \mathcal{MTL}$ against the models proposed in \citet{bhagavatula2019abductive}: a
majority baseline, supervised models (\textit{Infersent} and \textit{ESIM+ELMo}), as well as \textit{BERT$_{Large}$}. 
\citet{bhagavatula2019abductive} re-train the ESIM+ELMo and Infersent models on the $\mathcal{ART}$ dataset and fine-tuned the BERT model on the $\alpha$NLI task and report the results. 

We find that our \textbf{unsupervised} model with BERTScore ($LM_\mathcal{I}$ + BERTScore) outperforms (by $+9.28$ pp.\ and  $+1.28$ pp.) strong ESIM+ELMo and Infersent baseline models. Table \ref{tab:anli_examples} shows some examples of our generation model $LM_\mathcal{I}$ along with the obtained BERTScores. 

Unlike the unsupervised $LM_\mathcal{I}$ + BERTScore, our \textbf{supervised} $LM_\mathcal{I}+{\mathcal{MTL}}$ model also improves over the BERT$_{Large}$ baseline, by $+3.3$ pp. 
We can attribute the improvement to the model having been jointly trained to assess the similarity and dissimilarity of possible next events $O^{H_j}_{2}$ and observations ($O_2$) along with the \anli task.
One of the advantages of training our proposed
multi-task learning ($\mathcal{MTL}$) model, instead of directly feeding the possible next events $O^{H_j}_{2}$ as knowledge inputs is that it adds an explainable component to the model. One can view the generated next events $O^{H_j}_{2}$ as natural language rationales and our multi-task model explicitly chooses one of them. Hence, the multi-task framework makes the model more expressive. 
Finally,
we compare, for the $\mathcal{MTL}$ model, our embedded generation model
$LM_\mathcal{I}$ 
to pre-trained GPT-2 and COMET. Table \ref{tab:abnli} shows that 
\LMI + $\mathcal{MTL}$ yields better performance
compared to both \textit{COMET} +  $\mathcal{MTL}$ ($+3.1$ pp.) and \textit{GPT-2} + $\mathcal{MTL}$ ($+3.4$ pp.)
-- the intuitive reason being that the next events generated by \LMI are more helpful than events generated using pretrained GPT-2 and objects generated by COMET. 

Table \ref{tab:anli_examples} illustrates some examples where our $\mathcal{MTL}$ model not only chooses the correct hypothesis, but also a likely possible next event that is similar to the observation $O_2$. Interestingly, during training of \MTL  we initialize $w$ = 1, and after training the model we found the $w$ value had been adjusted to a range between
$0.85$ and $0.75$, which intuitively shows both the effectiveness of our \LMI-generated possible next events, and their similarity to the given observations $O_2$. 



\subsection{Case Study} Table \ref{tab:anli_examples} displays
possible next events, generated by our $LM_\mathcal{I}$ model -- along with the BERTscore measured between the possible next events $O^{H_j}_{2}$ and observation $O_2$. 
We see two different scenarios: (i) examples (a), (b) and (d) depicting the scenario where possible next events and observation pairs \textit{correctly} achieve higher BERTscores \footnote{BERTscore matches words in candidate and reference sentences by cosine similarity.}, (ii) example (c) depicting the scenario where an incorrect
possible next event and observation pair achieves higher BERTscores than the correct one. 
Intuitive reasons for these scenarios are, for example, for (a): there is a higher word overlap and semantic similarity between a correct next event and observation $O_2$, for example (b): there is higher semantic similarity; whereas for example (c): although there is a higher semantic dissimilarity, the word overlap between the wrong possible next event (\textit{``She started to feel sick."}) and the observation (\textit{``She felt much better afterwards."}) is much higher.

\section{Manual Evaluation}\label{sec:manual_eval}
Since the automatic scores only account for word-level similarity between observations and generated possible next events, we conduct a manual evaluation study, to assess
the quality of sentences generated by our \LMI model.

\paragraph{Annotation Study on \LMI generations.} The annotation was performed by three annotators with computational linguistic background. We provide each of the three annotators with observations, hypotheses and sentences, as produced by our \LMI model, for 50 randomly chosen instances from the $\alpha$NLI task. 
They obtain i) \textit{generated sentences for a next possible event} for
the \textit{correct} and \textit{incorrect hypothesis}, as well as ii) the \textit{sentence stating observation $O_2$}.  

We ask each annotator to rate the sentences according to four quality aspects as stated below.

\begin{description}
    \item[\textbf{Grammaticality:}] the sentence is i) grammatical, ii) not entirely grammatical but understandable, or iii) completely not understandable;\\[-7mm]
    \item[ \textbf{Redundancy:}] the sentence contains redundant or repeated information;\\[-7mm]
    \item[\textbf{Contradiction:}] 
     the sentence contains any pieces of information that are contradicting the given observation $O_2$ or not;\\[-7mm] 
    \item[\textbf{Relevance:}] the possible next event is i) relevant, ii) partially relevant, or iii) not relevant. 
\end{description}

For each aspect, they are asked to judge the sentence generated for the correct hypothesis\footnote{The correct hypothesis was marked for the annotation.}. Only for \textbf{Contradiction}, they are asked to judge both sentences, for correct and the incorrect hypotheses. 

\begin{figure}
    \centering
    \includegraphics[scale=1.0,height=4cm]{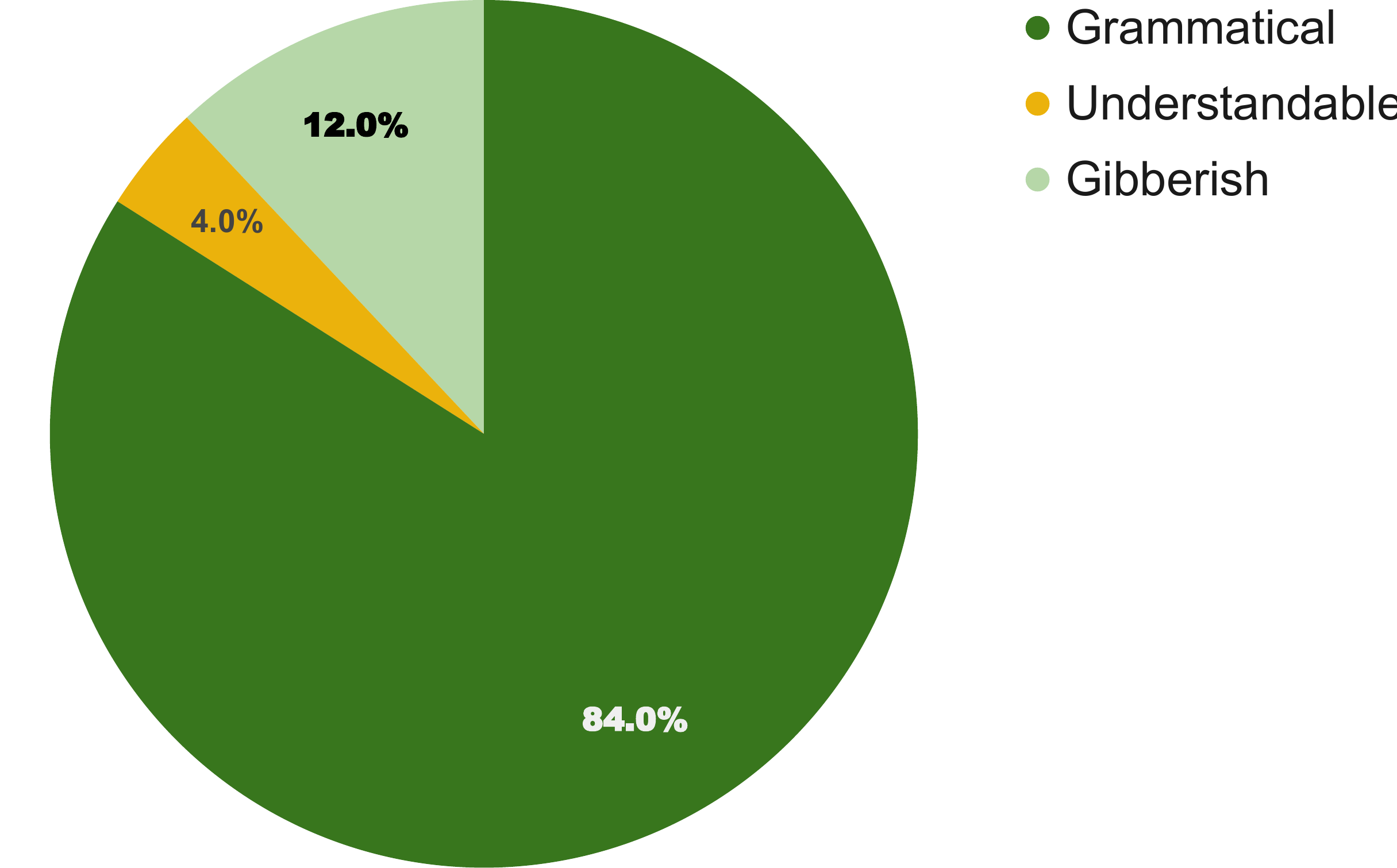}
    \caption{Human evaluation of the \textit{grammaticality} of generated sentences: ratio of i) grammatical, ii) not entirely grammatical but understandable, iii) completely not understandable sentences.}
    \label{fig:grammar}
\end{figure}

\begin{figure}
    \centering
    \includegraphics[scale=1.0,height=4cm]{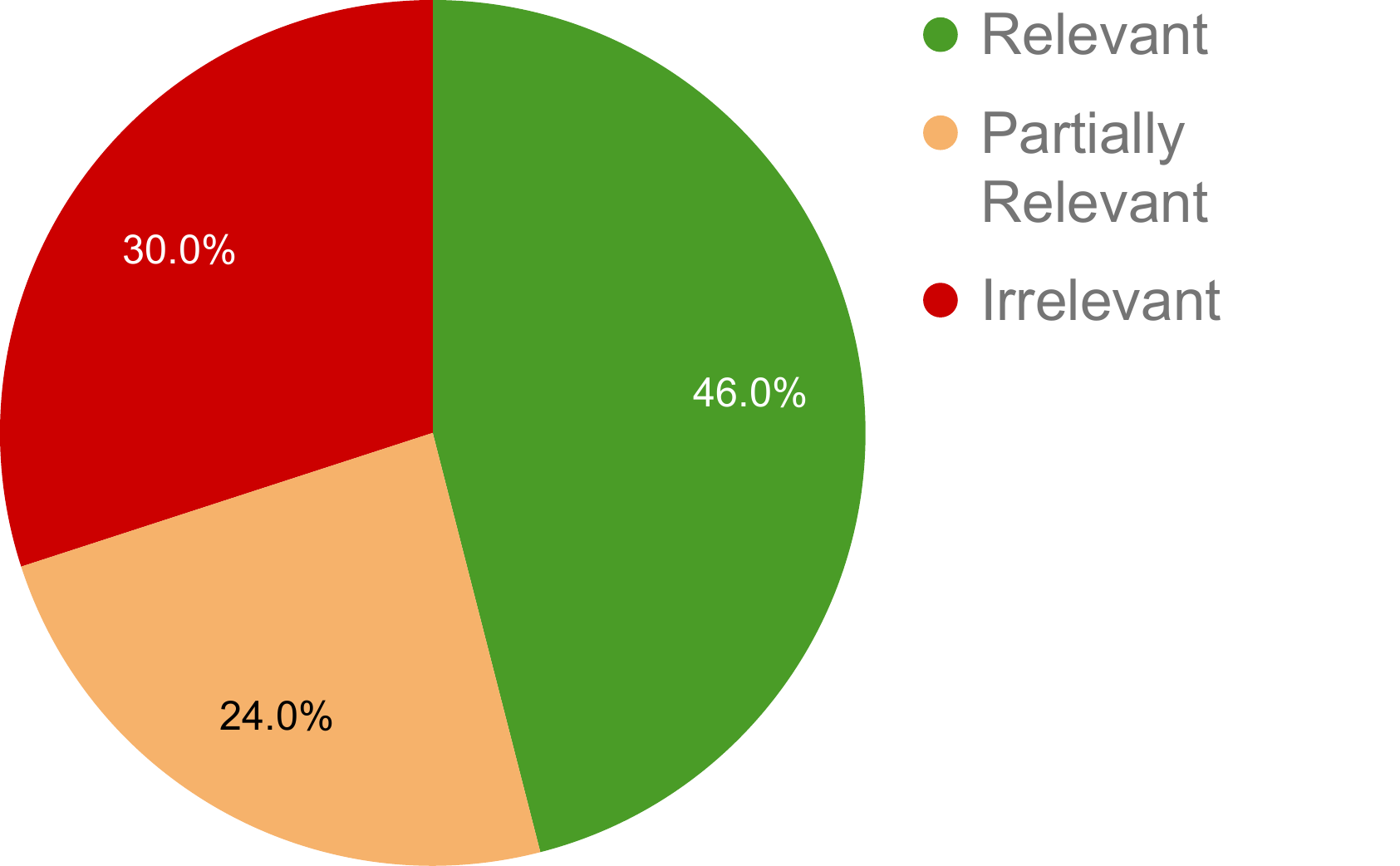}
    \caption{Human evaluation of the \textit{Relevance} of generated sentences for possible next events.}
    \label{fig:relevancy}
\end{figure}

\begin{figure}
    \centering
    \includegraphics[scale=1.0,height=4.5cm]{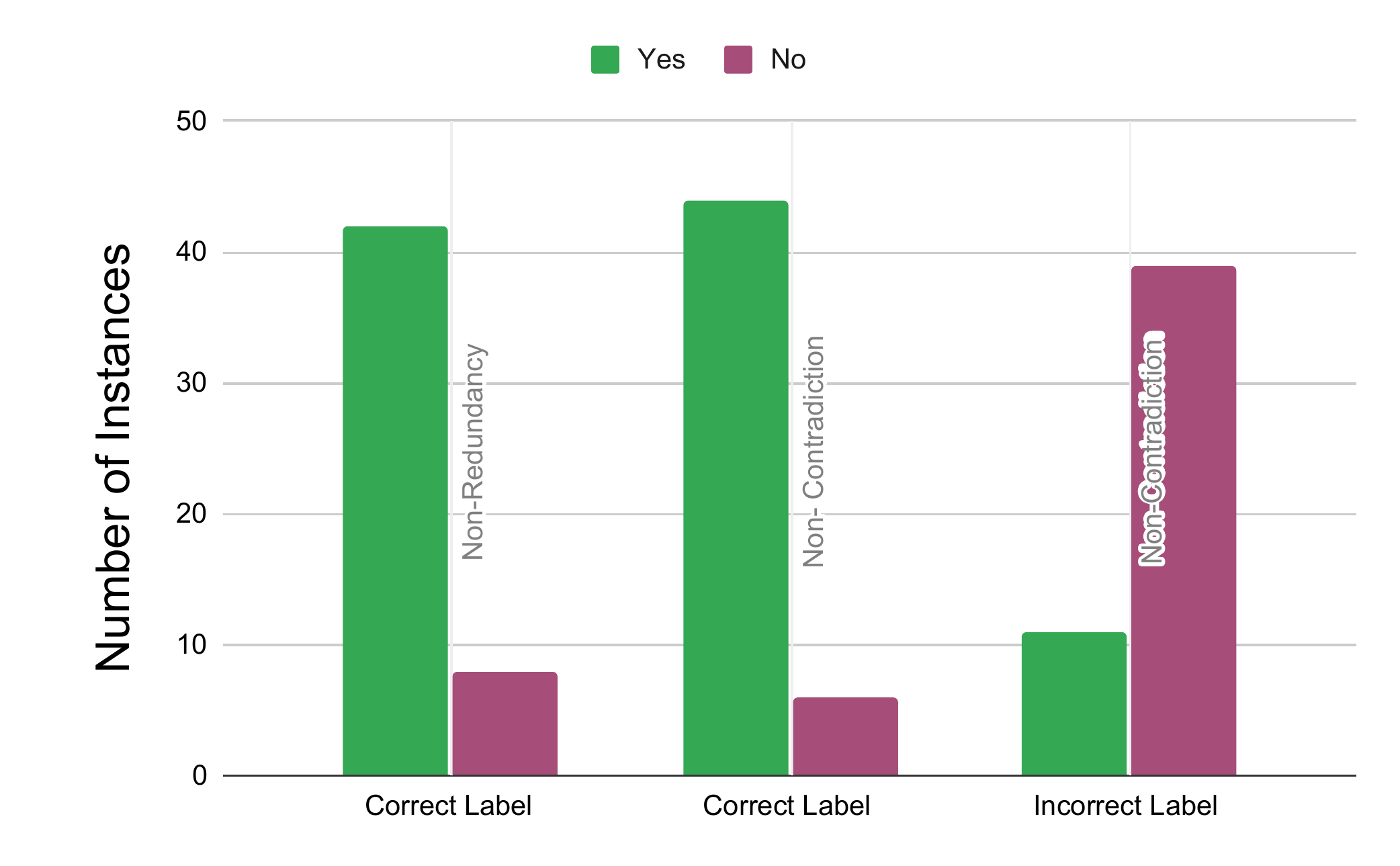}
    \caption{Human evaluation of \textit{Redundancy} and \textit{Contradiction} of generations for possible next events.}
    \label{fig:redundancyandcontradiction}
\end{figure}

\paragraph{Results and Discussion.} Figures \ref{fig:grammar}, \ref{fig:redundancyandcontradiction}, and \ref{fig:relevancy} present the results of manual evaluations of the generation quality, according to the different criteria described above.

For measuring inter-annotator agreement, we computed Krippendorff’s $\alpha$  \cite{Hayes2007AnsweringTC} for \textit{Grammaticality} and \textit{Relevance}, as it is suited for ordinal values, and Cohen's Kappa $\kappa$ for \textit{Redundancy} and \textit{Contradiction}. 
We found $\alpha$ values are $0.587$ and $0.462$ for \textit{Grammaticality} and \textit{Relevance}, respectively (moderate agreement) and $\kappa$ values $0.61$ and $0.74$ 
for \textit{Redundancy} and \textit{Contradiction} (substantial agreement). 
We aggregated the annotations from the three annotators using majority vote.
Figure \ref{fig:grammar} shows that the majority of sentences ($96$\%) are grammatical or understandable. Figure \ref{fig:redundancyandcontradiction} shows that most sentences for correct labels are non-redundant ($84$\%) and non-contradictory ($88$\%), whereas for incorrect labels 39 instances are found to be contradictory with the observation $O_2$ ($78$\%).  
The manual evaluation supports our hypothesis that the generated sentences for correct labels should be more similar (less contradictory) compared to the sentences generated for incorrect labels. Figure \ref{fig:relevancy} shows the ratio of sentences considered by humans as relevant, partially relevant, and irrelevant. The results show that $46$\% of cases are relevant (based on majority agreement) and $24$\% of cases are partially relevant. This yields that the generated sentences are (partially) relevant in most cases and thus should support abduction for both unsupervised ($LM_\mathcal{I}$ + BERTScore) and supervised ($LM_\mathcal{I}$ + $\mathcal{MTL}$) models. 

\paragraph{Impact of Reasoning types.} Finally, to better assess the performance of our model, we determine what \textit{types of reasoning} underly the abductive reasoning tasks in our data, and examine to what extent our models capture or not these reasoning types.  
We consider again the 50 instances that were annotated by our previous annotators
and manually classify them into different reasoning types. We broadly divided the data into 6 categories: (i) Motivation, (ii) Spatial-Temporal, (iii) Emotional, (iv) Negation, (v) Reaction, (vi) Situational fact.  The most frequent type was Emotional (10), most infrequent was Spatial (7). We ask a new annotator to annotate the reasoning types for these 50 instances. Considering the relevance and contradiction categories from the previous annotations
we determine that for Negation ($8$), Emotional ($10$), and Reaction ($8$) \textit{all} generated events for \textit{correct labels} are \textit{partially or fully relevant and non-contradictory}. 
An intuitive reason can be that we train our \LMI model to learn how different counterfactual hypothetical events emerging from a single premise can lead to the same or different outcomes through a series of events. Some counterfactual events ($s^{'}_2$) are negations of the  original event ($s_2$) in the TIMETRAVEL dataset. This may support the reasoning class Negation. 
For the other categories: Motivation, Spatial-temporal, and Situational fact,
we detect errors regarding (missing) \textit{Relevance} in $21$\%, $14$\% and $28$\% of cases,
respectively. 
Table \ref{tab:errorexample} illustrates an example from the class Situational Fact, where our generated next event is \textit{irrelevant} and \textit{redundant}.  

\begin{table}[t]
\centering

\scalebox{0.9}{
\begin{tabular}{l}
\textit{$O_1$: Jenna hit the weight hard in the gym.} \\
\textit{$O_2$: She took a cold bath in order to alleviate her pain.}\\
{\textit{$H_1$}: Her neck pain stopped because of this.} \\
{\textit{$H_2$}: \textbf{Jenna pulled a muscle lifting weights.}}\\
\textit{$O^{H_1}_{2}$: She decided to take a break .}  \\
\textit{$O^{H_2}_{2}$: Jenna lost weight in the gym. } \\
\end{tabular}}

\caption{Error Analysis: An example of generated possible next event $O^{H_j}_{2}$ from  Situational Fact category.}
\label{tab:errorexample}
\end{table}

\if false
Table \ref{tab:anli_examples} presents three examples with human evaluations. Interestingly, the third example indicating that an automatic evaluation like BERTScore might be insufficient in estimating the similarity and dissimilarity between events.
\fi

\section{Related Work}
\paragraph{Commonsense Reasoning.}  
There is 
growing interest in this research
field, which 
led to the creation of several new resources on commonsense reasoning, in form of both \textit{datasets}, such as SocialIQA \cite{sap-etal-2019-social}, CommonsenseQA \cite{talmor-etal-2019-commonsenseqa}, CosmosQA \cite{huang-etal-2019-cosmos} and \textit{knowledge bases},  e.g. ConceptNet \cite{speer2017conceptnet}, ATOMIC \cite{sapatomic}, or Event2Mind \cite{rashkin-etal-2018-event2mind}. 
Recently, many works proposed to utilize external \textit{static} knowledge graphs (KGs) to address the bottleneck of obtaining relevant commonsense knowledge. \citet{lin2019kagnet} proposed to utilize knowledge graph embeddings to rank and select relevant knowledge triples or paths. \citet{paul-frank-2019-ranking} proposed to extract subgraphs from KGs using graph-based ranking methods and further \citet{pauletal:2020} adopted the graph-based ranking method and proposed to \textit{dynamically} extend the KG to combat sparsity. 
In concurrent work, \citet{paulfrankcoins} introduced a method to dynamically generate contextually relevant knowledge that guides a  model while performing the narrative story completion task.

Both hypothetical reasoning and abductive reasoning are understudied problems in NLP. Recently, \citet{tandon-etal-2019-wiqa} proposed a first large-scale dataset of \textit{``What if..."} questions over procedural text. They introduced the dataset to study the effect of perturbations in procedural text. Related to our work, \citet{qin-counterfactual} investigated the capabilities of state-of-the-art LMs to rewrite stories with counterfactual reasoning. In our work we utilize this dataset to model how to generate possible next events emerging from different hypothetical and counterfactual events.  \citet{mostafazadeh-etal-2016-corpus} designed the narrative cloze task, a task to choose the correct ending of a story.\footnote{Their dataset, ROCStories, was later extended in \citet{qin-counterfactual} and \citet{bhagavatula2019abductive}.} Conversely, \citet{bhagavatula2019abductive} proposed a task that requires reasoning about plausible explanations for narrative omissions. Our research touches on the issue of hypothetical reasoning about alternative situations.  We found that making language models learn how different hypothetical events can evolve from a premise and result in similar or different future events forming from a premise and how these events can result in similar or different future events helps models to perform better in abduction.

\paragraph{Explainability.} Despite the success of large pre-trained language models, 
recent studies have raised some critical points such as:
high accuracy scores do not necessarily reflect understanding \cite{min-etal-2019-compositional}, large pretrained models may exploit superficial clues and annotation artifacts \cite{gururangan2018annotation, kavumba-etal-2019-choosing}. Therefore, the ability of models to generate 
explanations has become desirable, as this enhances 
interpretability. Recently,
there has been substantial effort to build datasets with natural language explanations \cite{NIPS2018_8163,Park2018MultimodalEJ, thayaparan2020survey}. There have also been numerous research works proposing 
models that are interpretable or explainable \cite{rajani-etal-2019-explain, atanasova-etal-2020-generating, latcinnik2020explaining, wiegreffe-marasovic-2021-review}.
Our work sheds light in this direction, as our ${\mathcal{MTL}}$ model not only predicts the plausible hypothesis $H_j$ but also generates possible next events $O^{H_j}_2$ and chooses the one that is closer to the given context,
thereby making our model more expressive. 

\paragraph{Abductive Reasoning.} There has been longstanding work on theories of abductive reasoning \citep{peirce1, charles2, charles3, kuipers1992naive, kuipers2013instrumentalism}. 
Researchers have applied various frameworks, some focused on 
pure logical frameworks \citep{pople1973mechanization, kakas1992abductive}, some on probabilistic frameworks  
\citep{pearl1988probabilistic}, and others
on Markov Logics \cite{singla2011abductive}. 
Recently, moving away from logic-based abductive reasoning, \citet{bhagavatula2019abductive} proposed to study 
language-based abductive reasoning. They introduced two tasks: \textit{Abductive Natural Language Inference ($\alpha$NLI)} and \textit{Generation ($\alpha$NLG)}. They establish baseline performance based on state-of-the-art language models and make use of inferential structured knowledge from ATOMIC \cite{sapatomic} as background knowledge. \citet{rankingforabductive} proposed to use a learning-to-rank framework to address the abductive reasoning task. \citet{ji-etal-2020-language} proposed a model GRF that enables pre-trained models (GPT-2) with dynamic multi-hop reasoning on multi-relational paths extracted from the external ConceptNet commonsense knowledge graph for the $\alpha$NLG task. \citet{paul-frank-2020-social} have proposed a multi-head knowledge attention method to incorporate commonsense knowledge to tackle the $\alpha$NLI task. Unlike our previous work in \citet{paul-frank-2020-social},
which focused on leveraging structured knowledge, in this work, we focus on learning about what will happen next from different counterfactual situations in a story context through language model fine-tuning.
Specifically,
we study the impact of such forward inference on the $\alpha$NLI task in a multi-task learning framework and show how it can improve performance over a strong BERT model.


\section{Conclusion}
We have introduced a novel method
for addressing the abductive 
reasoning task by explicitly learning what events
could follow other events in a hypothetical scenario, and learning to generate such events, conditioned on a premise or hypothesis. 
We show how a language model -- fine-tuned for this capability on a suitable narrative dataset -- can be leveraged to support abductive reasoning in the $\alpha$NLI tasks, in two settings: an unsupervised setting in combination with \textit{BertScore}, to select the proper hypothesis, and a supervised setting in a $\mathcal{MTL}$ setting. 

The relatively strong performance of our proposed models demonstrates that learning to choose from generated hypothetical
next events the one that is most similar to the observation, supports the prediction of the most plausible hypothesis. Our experiments show that our unsupervised $LM_\mathcal{I}+$\textit{BERTScore} model outperforms some of the strong supervised baseline systems on $\alpha$NLI. Our research thus offers new perspectives for training generative models in different ways for various complex reasoning tasks.

\section*{Acknowledgements}

This work has been supported by the German Research Foundation as part of the Research
Training Group “Adaptive Preparation of Information from Heterogeneous Sources” (AIPHES)
under grant No.\ GRK 1994/1. We thank our annotators for their valuable annotations. We also thank NVIDIA Corporation for donating GPUs used in this research.

\bibliography{acl2021}
\bibliographystyle{acl_natbib}

\end{document}